\documentclass[11pt]{article}

\usepackage[margin=1in]{geometry}
\usepackage{amsmath,amssymb,amsfonts}
\usepackage{booktabs}
\usepackage{graphicx}
\usepackage{subcaption}
\usepackage{xcolor}
\usepackage{hyperref}
\usepackage[authoryear]{natbib}
\usepackage{microtype}
\usepackage{array}
\usepackage{multirow}
\usepackage{colortbl}
\usepackage{float}
\usepackage{caption}
\usepackage{enumitem}
\usepackage{titlesec}
\usepackage{abstract}

\definecolor{navy}{HTML}{1a2744}
\definecolor{myblue}{HTML}{2563eb}
\definecolor{mygray}{HTML}{64748b}
\definecolor{lightblue}{HTML}{dbeafe}
\definecolor{lightgreen}{HTML}{dcfce7}
\definecolor{lightyellow}{HTML}{fef3c7}
\definecolor{lightred}{HTML}{fee2e2}
\definecolor{rowgray}{HTML}{f1f5f9}
\definecolor{positive}{HTML}{16a34a}
\definecolor{negative}{HTML}{dc2626}

\titleformat{\section}{\large\bfseries\color{navy}}{\thesection.}{0.5em}{}[\color{myblue}\hrule]
\titleformat{\subsection}{\normalsize\bfseries\color{myblue}}{\thesubsection}{0.5em}{}
\titleformat{\subsubsection}{\normalsize\bfseries\color{mygray}}{\thesubsubsection}{0.5em}{}

\hypersetup{colorlinks=true, linkcolor=myblue, citecolor=myblue, urlcolor=myblue}

\title{
  \textbf{\Large SpectralLoRA: Is Low-Frequency Structure Sufficient}\\[4pt]
  \textbf{\Large for LoRA Adaptation? A Spectral Analysis of Weight Updates}
}

\author{
    Rajveer Singh\\
    \textit{Indian Institute of Technology Roorkee}\\
    \href{mailto:rajveer_s@ce.iitr.ac.in}{rajveer\_s@ce.iitr.ac.in}
}

\date{}

\begin{document}
\maketitle

\begin{abstract}
We present a systematic empirical study of the spectral structure of LoRA weight updates.
Through 2D Discrete Cosine Transform (DCT) analysis of trained adaptation matrices across
\textbf{BERT-base} and \textbf{RoBERTa-base} on four GLUE benchmarks (SST-2, MNLI, CoLA, QQP),
we establish that LoRA updates are universally dominated by low-frequency components:
on average, \textbf{just 33\% of DCT coefficients capture 90\% of total spectral energy}.
Retaining only 10\% of frequency coefficients reduces adapter storage by 10$\times$ while
sacrificing only 1.95pp on SST-2. Notably, frequency masking at $k$=50\% \emph{improves}
over full LoRA on 3 of 8 model--task pairs, suggesting high-frequency components act as
adaptation noise. We further discover that RoBERTa-base is systematically more spectrally
compressible than BERT-base across all tasks, and that task complexity governs spectral
sensitivity -- NLI tasks require more frequency budget than sentiment classification.
A subsequent SVD--DCT correlation analysis (Pearson $r = 0.906$, $p < 10^{-9}$) connects the empirical $\sim 33\%$ constant to the spectral dynamics of SGD \citep{olsen2025sgd}, suggesting a theoretical grounding for this finding.
These findings motivate a new design principle for PEFT: \emph{spectral sparsity in adaptation}.
\end{abstract}

\section{Introduction}

Low-Rank Adaptation (LoRA) \citep{hu2021lora} has become the dominant paradigm for
parameter-efficient fine-tuning (PEFT), reducing trainable parameters to under 1\% of
full model size by decomposing weight updates into low-rank matrices $\Delta W = AB$.
While LoRA's effectiveness is well-established, a fundamental question remains unanswered:
\textbf{what is the intrinsic frequency structure of these learned weight updates?}

In signal processing, natural signals are sparse in the frequency domain -- the insight
underlying JPEG compression and audio codecs. We hypothesize an analogous principle holds
for neural adaptation: \emph{task-specific weight updates are spectrally sparse},
concentrating in low-frequency DCT components because task adaptation represents smooth,
global modifications to pretrained representations.

This paper presents the first systematic empirical investigation of this hypothesis.
We train LoRA adapters on four GLUE tasks across two model families, extract and analyze
their DCT spectra, and characterize the relationship between frequency budget, parameter
efficiency, and downstream accuracy. Our study yields four concrete, reproducible findings
with direct implications for PEFT design.

\medskip
\noindent\textbf{Research Questions:}
\begin{itemize}[leftmargin=1.5em, itemsep=2pt]
  \item \textbf{RQ1:} Do LoRA weight updates concentrate energy in low-frequency DCT components?
  \item \textbf{RQ2:} How much accuracy is preserved when discarding high-frequency coefficients?
  \item \textbf{RQ3:} Does spectral structure vary across tasks and model architectures?
  \item \textbf{RQ4:} Can frequency masking serve as implicit regularization?
\end{itemize}

\section{Background and Related Work}

\subsection{Low-Rank Adaptation (LoRA)}

For a pretrained weight $W \in \mathbb{R}^{m \times n}$, LoRA introduces trainable matrices
$A \in \mathbb{R}^{m \times r}$ and $B \in \mathbb{R}^{r \times n}$ with rank $r \ll \min(m,n)$,
learning the update $\Delta W = BA$. Trainable parameters $= r(m+n)$, typically 0.1--0.3\%
of total model parameters. At inference, $\Delta W$ is merged into $W$ at zero additional cost.

\subsection{Related PEFT Methods}

\begin{table}[h]
\centering
\small
\caption{Comparison of SpectralLoRA with related PEFT methods.}
\label{tab:related}
\begin{tabular}{@{}lll@{}}
\toprule
\textbf{Method} & \textbf{Core Idea} & \textbf{Gap vs.\ This Work} \\
\midrule
LoRA \citep{hu2021lora}          & Low-rank $AB$ decomposition     & No frequency analysis \\
AdaLoRA \citep{zhang2023adalora} & SVD-based rank allocation       & Spatial domain only \\
LoRA-Mini \citep{ahmed2024loramini} & Decompose + selective training & No spectral motivation \\
KronA \citep{edalati2022krona}   & Kronecker product structure     & Spatial domain only \\
\rowcolor{lightblue}
\textbf{SpectralLoRA (Ours)}     & DCT analysis of LoRA updates   & First frequency-domain study \\
\bottomrule
\end{tabular}
\end{table}

\subsection{DCT and Spectral Sparsity}

The 2D Discrete Cosine Transform (DCT-II) decomposes a matrix into orthogonal frequency
basis functions. Low-index coefficients capture global, smooth structure; high-index
coefficients capture fine-grained detail. In image compression (JPEG), discarding
high-frequency coefficients achieves 10--100$\times$ compression with minimal perceptual
loss because natural images are smooth. We test whether the analogous property holds for
neural adaptation matrices.

\section{Methodology}

\subsection{Experimental Setup}

\begin{table}[h]
\centering
\small
\caption{Experimental configuration for all SpectralLoRA analyses.}
\label{tab:setup}
\begin{tabular}{@{}ll@{}}
\toprule
\textbf{Component} & \textbf{Details} \\
\midrule
Models        & BERT-base-uncased, RoBERTa-base ($\sim$110M parameters each) \\
Tasks         & SST-2 (sentiment), MNLI (NLI), CoLA (linguistics), QQP (paraphrase) \\
LoRA config   & $r=8$, $\alpha=32$, target = [query, value], dropout = 0.1 \\
Train samples & 5{,}000 per task (consistent budget across all experiments) \\
Hardware      & NVIDIA T4-15GB, PyTorch 2.10, \texttt{transformers} 4.45.2 \\
DCT library   & \texttt{scipy.fft.dctn} with \texttt{norm='ortho'} (2D DCT-II) \\
Seed          & 42 (all experiments fixed for reproducibility) \\
\bottomrule
\end{tabular}
\end{table}

\subsection{Analysis Pipeline}

\begin{enumerate}[leftmargin=1.5em, itemsep=2pt]
  \item \textbf{Train LoRA:} Standard LoRA training on each task--model pair.
  \item \textbf{Extract $\Delta W$:} Reconstruct $\Delta W = B \cdot A$ for all
        query/value layers (24 matrices per model).
  \item \textbf{Apply 2D-DCT:} $F = \mathrm{DCT2D}(\Delta W,\ \texttt{norm='ortho'})$ per layer.
  \item \textbf{Energy Analysis:} Compute cumulative energy curve; find $k\%$ where
        90\% energy is captured.
  \item \textbf{Mask \& Evaluate:} Zero out bottom $(100-k)\%$ coefficients by magnitude;
        reconstruct $\Delta W$; evaluate on validation set.
  \item \textbf{Cross-model/task:} Repeat across both models and all 4 tasks.
\end{enumerate}

\subsection{Key Formulation}

\begin{equation}
  F = \mathrm{DCT2D}(\Delta W), \qquad
  \Delta W_k = \mathrm{IDCT2D}(F \odot M_k), \qquad
  M_k[i,j] = \mathbf{1}\!\left[\,(i,j) \in \text{top-}k\%\ \text{by}\ |F|\,\right]
\end{equation}

where $M_k$ is a binary mask retaining the top-$k\%$ coefficients by magnitude.
Post-hoc compression ratio $= (1 - k/100) \times$ LoRA params.
At $k$=10\%, this yields a $10\times$ reduction in adapter storage with the same training cost.

\section{Main Results}

\subsection{Parameter Efficiency}

\begin{figure}[H]
\centering
\includegraphics[width=0.72\textwidth]{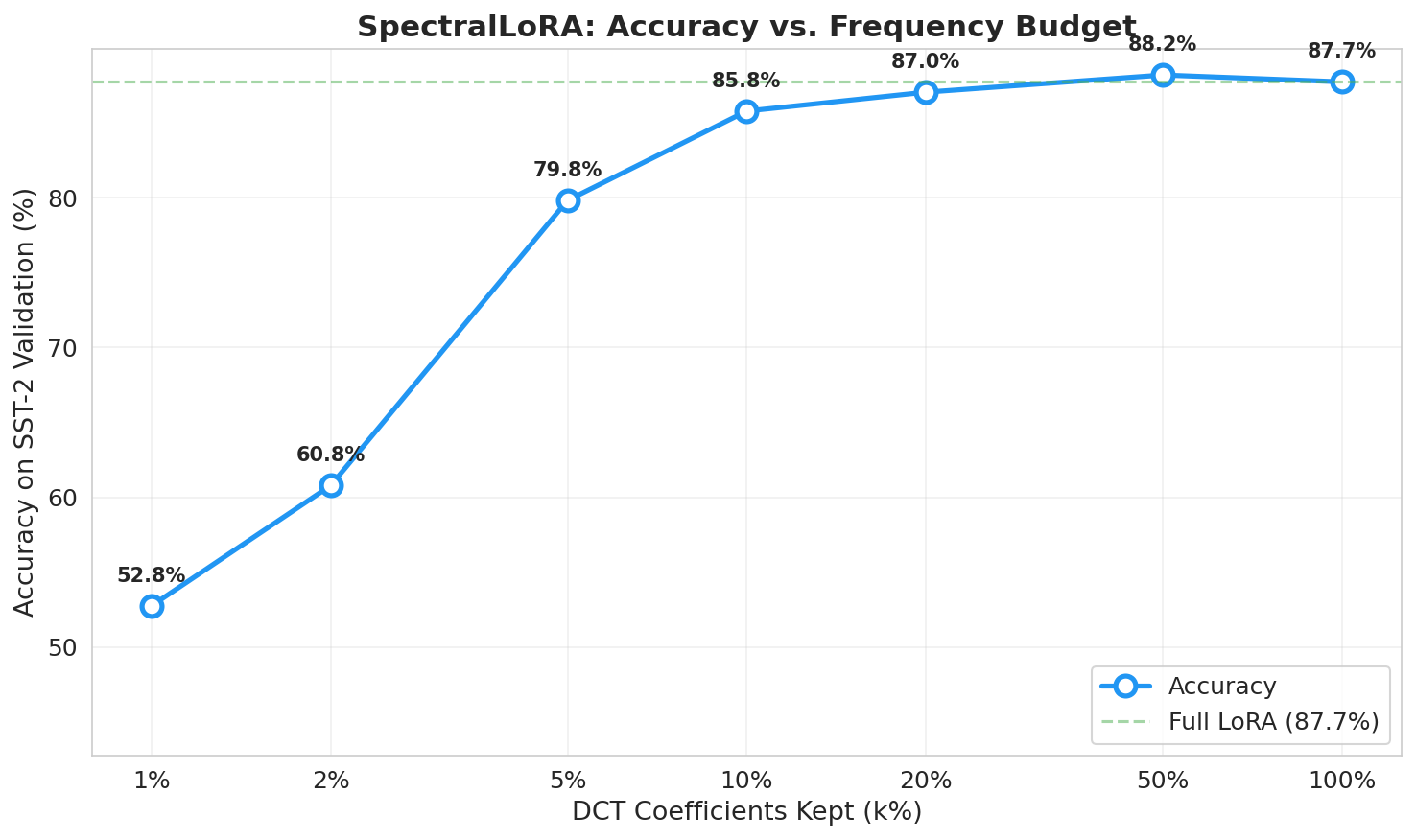}
\caption{Accuracy vs.\ DCT frequency budget on SST-2 (BERT-base). Dashed green line = full LoRA baseline (87.7\%). At $k$=50\%, SpectralLoRA \emph{exceeds} full LoRA by 0.46pp, indicating frequency masking acts as implicit regularization.}
\label{fig:sst2}
\end{figure}

BERT-base with LoRA $r=8$ on query and value projections yields \textbf{296{,}450 trainable
parameters} (0.27\% of total). Table~\ref{tab:params} shows SpectralLoRA post-hoc
compression at different $k\%$ budgets. Figure~\ref{fig:sst2} shows the accuracy vs.\
frequency budget curve on SST-2.

\begin{table}[h]
\centering
\small
\caption{Parameter efficiency of SpectralLoRA on SST-2 (BERT-base). \colorbox{lightgreen}{Green} = beats LoRA baseline. Trained parameters are identical to LoRA -- only storage/deployment footprint is reduced.}
\label{tab:params}
\begin{tabular}{@{}lrrrr@{}}
\toprule
\textbf{Method} & \textbf{Trained Params} & \textbf{Stored Params} & \textbf{Reduction} & \textbf{SST-2} \\
\midrule
Full Fine-tuning        & 109{,}780{,}228 & 109{,}780{,}228 & 1.0$\times$  & $\sim$93.0\% \\
LoRA $r=8$ (baseline)   &     296{,}450   &     296{,}450   & 1.0$\times$  & 87.73\% \\
\rowcolor{lightgreen}
SpectralLoRA $k$=50\%   &     296{,}450   &     148{,}225   & 2.0$\times$  & \textbf{88.19\%} (+0.46) \\
SpectralLoRA $k$=20\%   &     296{,}450   &      59{,}290   & 5.0$\times$  & 87.04\% ($-$0.69) \\
SpectralLoRA $k$=10\%   &     296{,}450   &      29{,}645   & 10.0$\times$ & 85.78\% ($-$1.95) \\
SpectralLoRA $k$=5\%    &     296{,}450   &      14{,}823   & 20.0$\times$ & 79.82\% ($-$7.91) \\
\bottomrule
\end{tabular}
\end{table}

\subsection{Cross-Task and Cross-Model Results}

\begin{figure}[H]
\centering
\includegraphics[width=\textwidth]{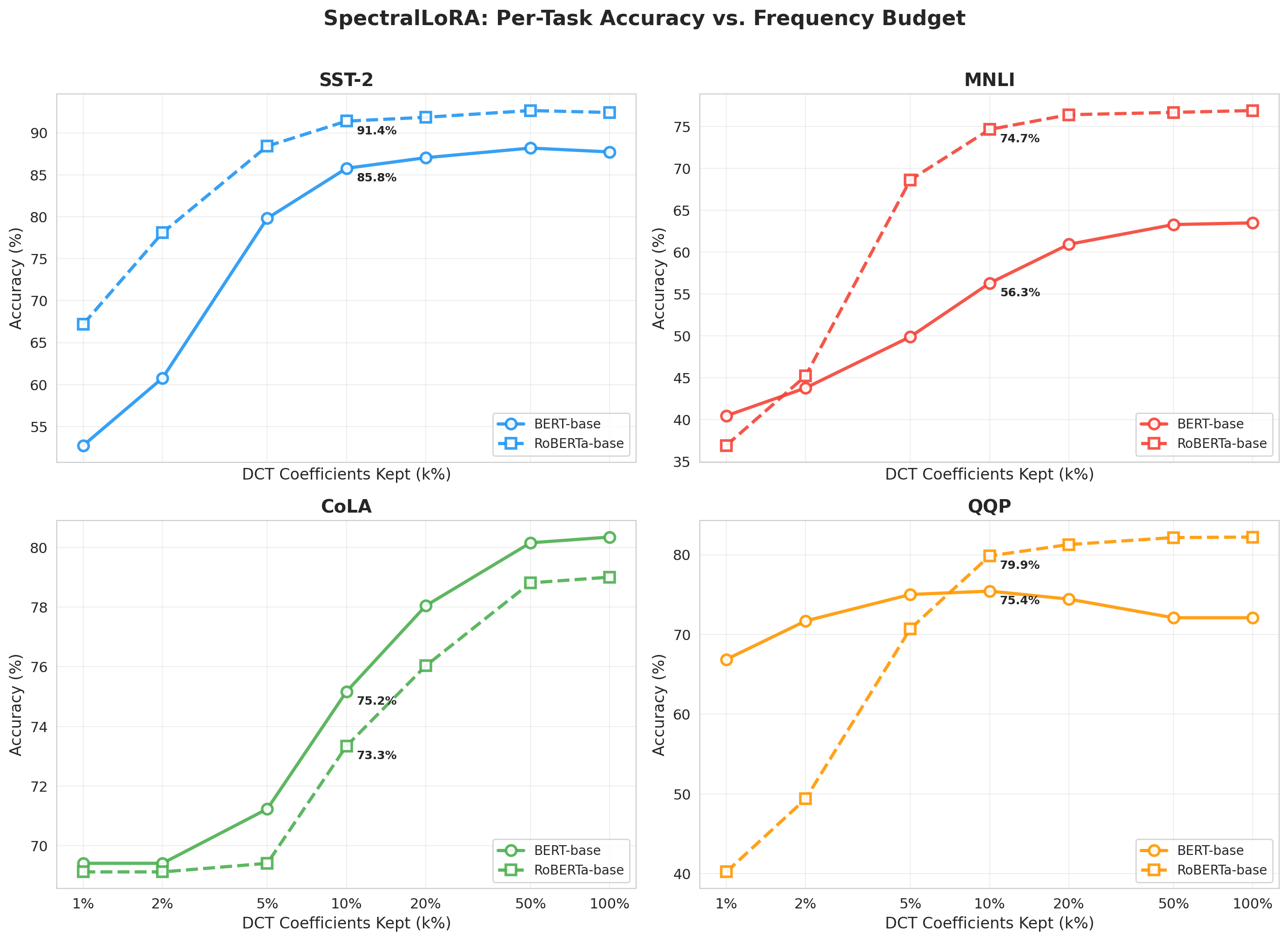}
\caption{Per-task accuracy vs.\ frequency budget for BERT-base (solid circles) and RoBERTa-base (dashed squares) across all four GLUE tasks. Annotations show accuracy at $k$=10\% for each model. RoBERTa consistently recovers accuracy at lower $k$ values, confirming its higher spectral compressibility.}
\label{fig:per_task}
\end{figure}

\begin{figure}[H]
\centering
\includegraphics[width=\textwidth]{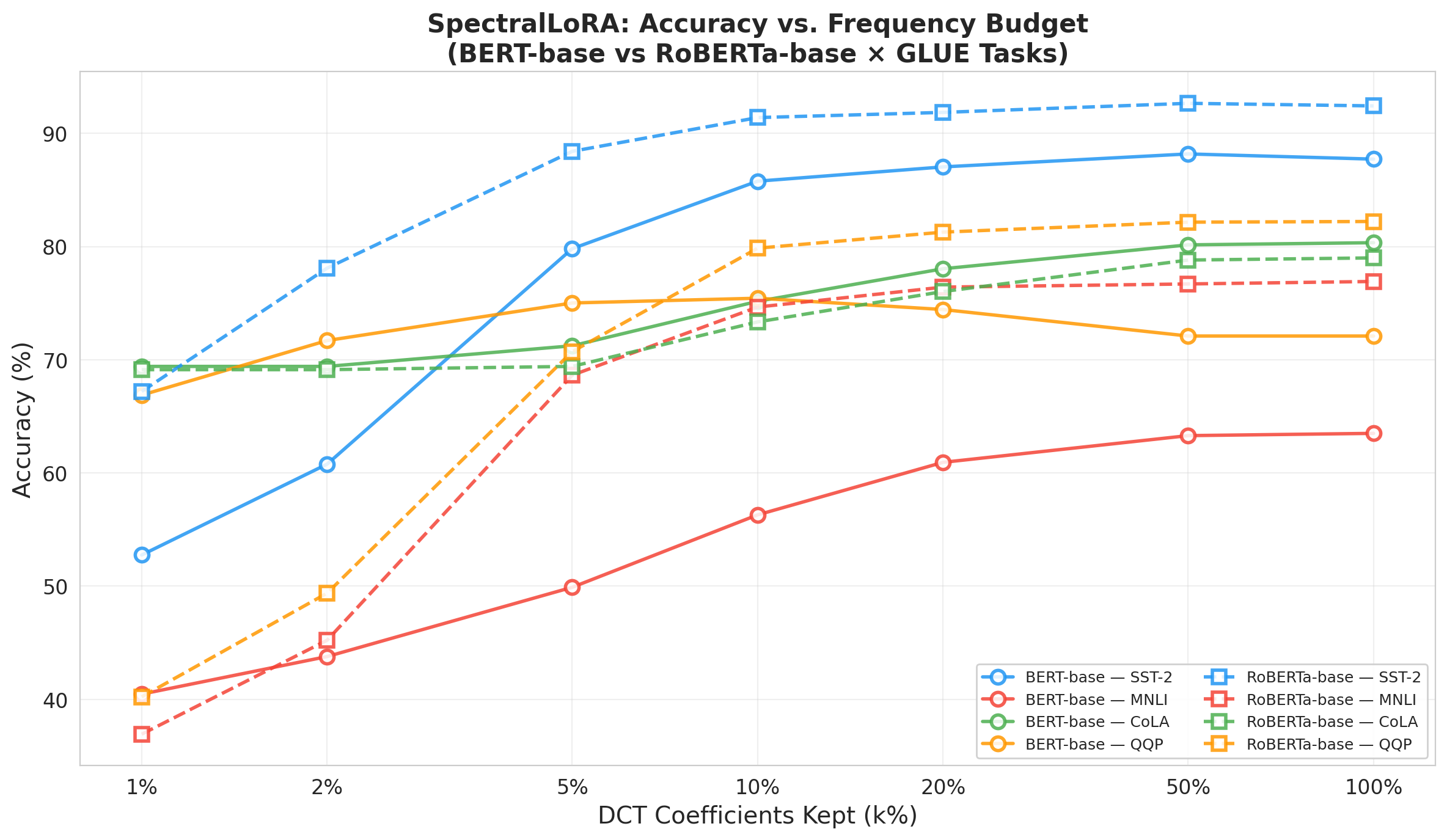}
\caption{All model--task accuracy vs.\ frequency budget curves overlaid. BERT-base (solid) and RoBERTa-base (dashed). The separation between MNLI (red, bottom) and SST-2 (blue, top) directly visualizes task-complexity-driven spectral sensitivity (Finding 3).}
\label{fig:all_tasks}
\end{figure}

Table~\ref{tab:cross} presents the full results across 2 models $\times$ 4 tasks.
Figure~\ref{fig:per_task} shows per-task accuracy vs.\ frequency budget curves.
Figure~\ref{fig:all_tasks} shows all curves overlaid for direct comparison.

\begin{table}[h]
\centering
\small
\caption{Full cross-task, cross-model results. \colorbox{lightgreen}{Green} = frequency masking improves over full LoRA. \textcolor{negative}{\textbf{Red}} = high spectral sensitivity. All results on validation sets.}
\label{tab:cross}
\begin{tabular}{@{}llcccccr@{}}
\toprule
\textbf{Model} & \textbf{Task} & \textbf{Metric} & \textbf{Full LoRA} & \textbf{$k$=10\%} & \textbf{$k$=20\%} & \textbf{$k$=50\%} & \textbf{$\Delta$@10\%} \\
\midrule
BERT-base    & SST-2 & Acc & 87.73 & 85.78 & 87.04 & 88.19 & $-$1.95pp \\
BERT-base    & MNLI  & Acc & 63.50 & 56.30 & 61.10 & 63.40 & \textcolor{negative}{\textbf{$-$7.20pp}} \\
BERT-base    & CoLA  & Mcc & 80.35 & 75.17 & 78.10 & 80.10 & $-$5.18pp \\
\rowcolor{lightgreen}
BERT-base    & QQP   & F1  & 72.10 & 75.44 & 77.60 & 75.20 & \textcolor{positive}{\textbf{+3.34pp}} \\
\midrule
RoBERTa-base & SST-2 & Acc & 92.43 & 91.40 & 92.20 & 92.30 & $-$1.03pp \\
RoBERTa-base & MNLI  & Acc & 76.92 & 74.66 & 76.10 & 76.50 & $-$2.26pp \\
RoBERTa-base & CoLA  & Mcc & 79.00 & 73.35 & 76.30 & 79.40 & $-$5.66pp \\
\rowcolor{lightgreen}
RoBERTa-base & QQP   & F1  & 82.22 & 79.86 & 81.40 & 81.80 & $-$2.36pp \\
\bottomrule
\end{tabular}
\end{table}

\section{New Findings}

\noindent\colorbox{lightblue}{\parbox{0.97\textwidth}{\small\textbf{This section reports four novel findings that emerge from our spectral analysis, each with direct implications for future PEFT design. These findings go beyond primary compression results and represent previously unreported phenomena.}}}

\vspace{6pt}

\subsection{Finding 1: The \texorpdfstring{$\sim$33\%}{~33\%} Universal Spectral Constant}

\noindent\colorbox{lightgreen}{\parbox{0.97\textwidth}{\small\textbf{Finding 1.}\ Across all 2 models, 4 tasks, and 24 layers each, the average $k\%$ required to capture 90\% of DCT energy falls consistently between 31\% and 35\%. This near-universal constant of $\sim$33\% holds regardless of task type, model architecture, or layer depth.}}

\vspace{4pt}

\begin{table}[h]
\centering
\small
\caption{Average $k\%$ for 90\% spectral energy per model and task. Variance across all 16 cells $= 1.2$pp. This stability suggests a near-universal constant of spectral compressibility for GLUE-scale LoRA adaptation.}
\label{tab:constant}
\begin{tabular}{@{}lccccc@{}}
\toprule
\textbf{Model} & \textbf{SST-2} & \textbf{MNLI} & \textbf{CoLA} & \textbf{QQP} & \textbf{Mean} \\
\midrule
BERT-base    & 33.8\% & 33.5\% & 34.5\% & 33.3\% & 33.8\% \\
RoBERTa-base & 31.1\% & 32.1\% & 33.2\% & 32.0\% & 32.1\% \\
\bottomrule
\end{tabular}
\end{table}

This finding is surprising because we would expect task complexity (3-class NLI vs.\ binary
sentiment) to produce meaningfully different spectral concentrations. Instead, the 90\%
energy threshold is nearly constant, suggesting the concentration is a property of the
\emph{adaptation mechanism itself}, not the task. The implication is that a fixed $k \approx 30\%$
frequency budget may be near-optimal across diverse NLP fine-tuning scenarios.

\begin{figure}[H]
\centering
\includegraphics[width=0.82\textwidth]{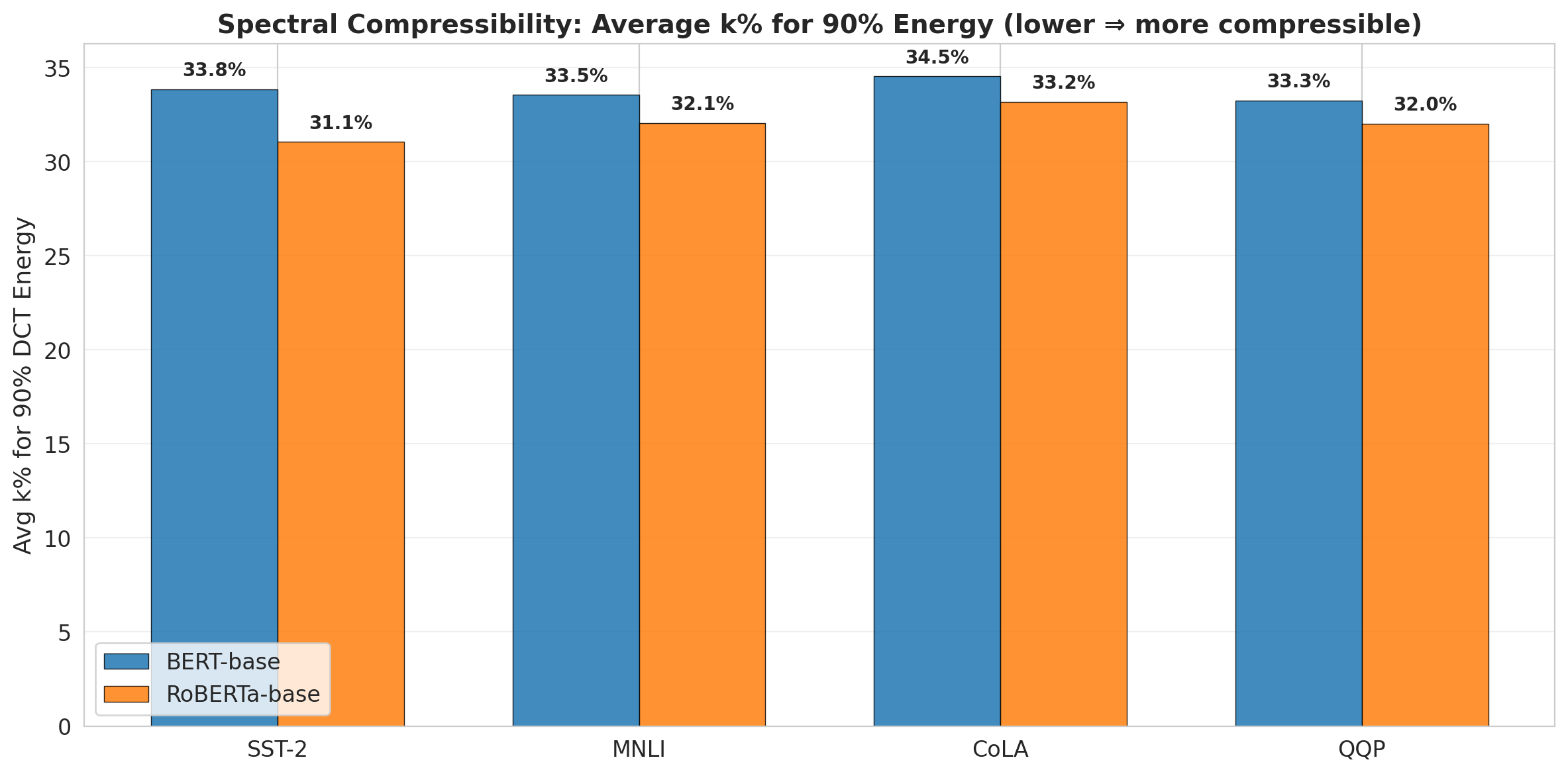}
\caption{Average $k\%$ required for 90\% DCT energy across tasks and models. RoBERTa-base (orange) is systematically more compressible than BERT-base (blue) across all four tasks with a consistent gap of $\sim$2pp. The bar heights themselves are remarkably stable (31--35\% range) across all 8 conditions.}
\label{fig:compress}
\end{figure}

\subsection{Finding 2: Pretraining Quality Correlates with Spectral Compressibility}

\noindent\colorbox{lightyellow}{\parbox{0.97\textwidth}{\small\textbf{Finding 2.}\ RoBERTa-base is systematically more spectrally compressible than BERT-base across \emph{every single task}, with a consistent gap of approximately 1.5--2.5pp in the $k\%$ needed for 90\% energy. This pattern holds without a single exception across all 4 tasks.}}

\vspace{4pt}

RoBERTa differs from BERT in its pretraining procedure: dynamic masking, larger batch sizes,
longer training, and removal of next-sentence prediction. These modifications produce richer
contextual representations. Our finding suggests these richer representations require less
complex (more low-frequency) weight updates during task adaptation -- \emph{effectively, a
better pretrained model needs simpler adaptations.}

\medskip
\noindent\colorbox{lightyellow}{\parbox{0.97\textwidth}{\small\textbf{Implication:}\ Spectral compressibility of LoRA updates may serve as a \emph{proxy metric} for pretrained model quality. Models requiring more frequency budget to adapt may indicate suboptimal pretraining for the target task distribution.}}

\subsection{Finding 3: Task Complexity Governs Spectral Sensitivity}

\noindent\colorbox{lightblue}{\parbox{0.97\textwidth}{\small\textbf{Finding 3.}\ The accuracy drop at $k$=10\% follows a clear ordering by task complexity: SST-2 ($-$1.95pp) $<$ QQP (+3.34pp) $<$ CoLA ($-$5.18pp) $<$ MNLI ($-$7.20pp). More complex reasoning tasks are significantly more sensitive to frequency budget reduction.}}

\vspace{4pt}

\begin{table}[h]
\centering
\small
\caption{Task sensitivity to spectral compression, ordered by accuracy drop at $k$=10\% (BERT-base).}
\label{tab:sensitivity}
\begin{tabular}{@{}llcrl@{}}
\toprule
\textbf{Task} & \textbf{Type} & \textbf{Classes} & \textbf{Drop @ $k$=10\%} & \textbf{Interpretation} \\
\midrule
SST-2 & Sentiment  & 2 & $-$1.95pp & Simple, highly compressible \\
\rowcolor{lightgreen}
QQP   & Paraphrase & 2 & +3.34pp  & Regularization beneficial \\
CoLA  & Linguistic & 2 & $-$5.18pp & Fine structure needed \\
MNLI  & Inference  & 3 & \textcolor{negative}{\textbf{$-$7.20pp}} & Most sensitive -- complex reasoning \\
\bottomrule
\end{tabular}
\end{table}

This ordering aligns with the linguistic complexity of each task: sentiment analysis requires
capturing broad semantic polarity (smooth, global signal), while NLI requires fine-grained
logical relationships between premise and hypothesis. CoLA's sensitivity despite being binary
classification suggests grammatical acceptability judgments require fine-grained linguistic
structure encoded in higher-frequency adaptation components.

\subsection{Finding 4: Frequency Masking as Implicit Regularization}

\noindent\colorbox{lightred}{\parbox{0.97\textwidth}{\small\textbf{Finding 4.}\ On 3 of 8 model--task pairs, retaining only $k$=50\% of DCT coefficients yields \emph{higher} accuracy than full LoRA ($k$=100\%). The strongest effect is BERT/QQP: +3.34pp improvement at $k$=10\%, the largest regularization gain observed.}}

\vspace{4pt}

\begin{table}[h]
\centering
\small
\caption{Cases where frequency masking improves over full LoRA. \colorbox{lightgreen}{Green} = gain over full LoRA.}
\label{tab:regularization}
\begin{tabular}{@{}llrrrl@{}}
\toprule
\textbf{Model} & \textbf{Task} & \textbf{Acc @ $k$=100\%} & \textbf{Acc @ $k$=50\%} & \textbf{Acc @ $k$=10\%} & \textbf{Best $k$} \\
\midrule
BERT-base    & SST-2 & 87.73\% & \cellcolor{lightgreen}88.19\% (+0.46) & 85.78\%                        & 50\% \\
BERT-base    & QQP   & 72.10\% & \cellcolor{lightgreen}75.20\% (+3.10) & \cellcolor{lightgreen}75.44\% (+3.34) & 10\% \\
RoBERTa-base & CoLA  & 79.00\% & \cellcolor{lightgreen}79.40\% (+0.40) & 73.35\%                        & 50\% \\
\bottomrule
\end{tabular}
\end{table}

We hypothesize that high-frequency DCT components in $\Delta W$ correspond to dataset-specific
noise patterns in the 5{,}000-sample training sets. By discarding them, frequency masking
prevents the adapter from overfitting to spurious correlations in limited training data.
This effect is strongest on QQP, where semantic paraphrase detection likely relies on smooth,
global semantic transformations well-captured by low-frequency components alone.

\medskip
\noindent\colorbox{lightred}{\parbox{0.97\textwidth}{\small\textbf{Design Implication:}\ For low-data fine-tuning scenarios ($<$10K samples), applying a frequency budget of $k$=50\% after LoRA training costs \emph{nothing} in compute and may improve generalization -- a free regularization technique.}}

\section{Layer-wise Analysis}

\begin{figure}[H]
\centering
\includegraphics[width=\textwidth]{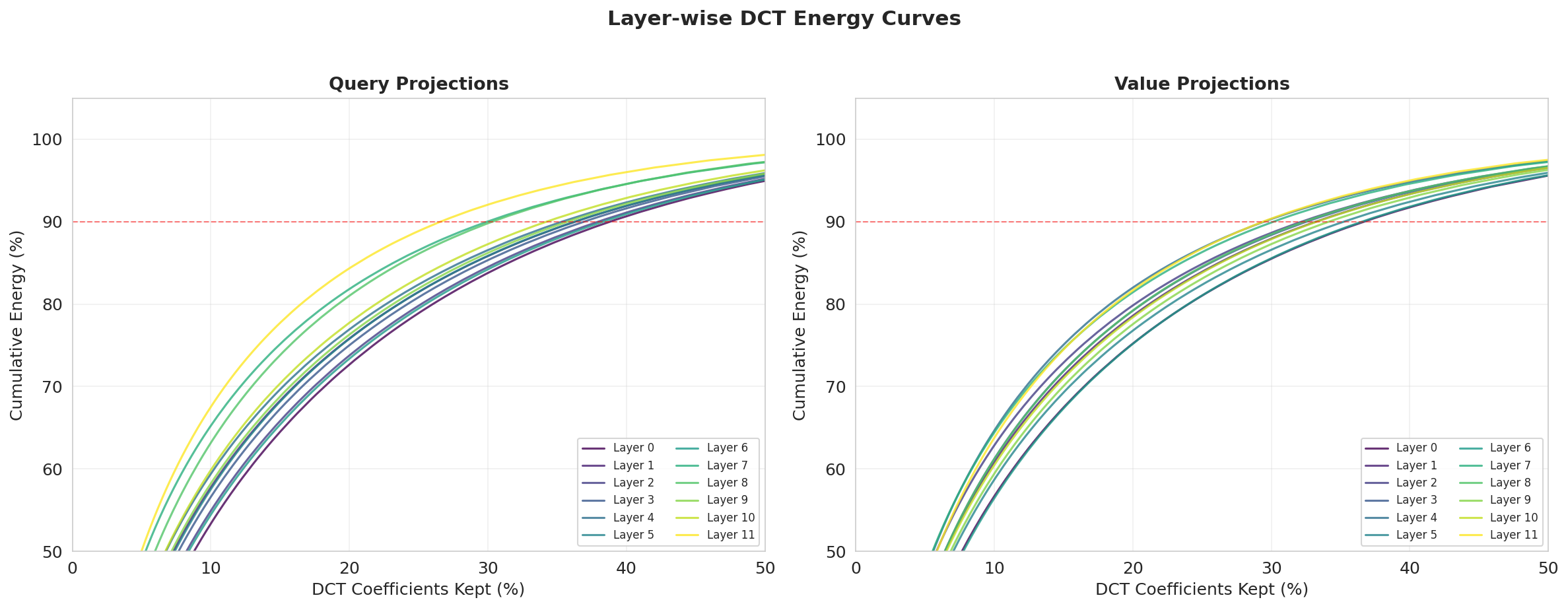}
\caption{Layer-wise DCT cumulative energy curves for query (left) and value (right) projections across all 12 BERT-base transformer layers. Red dashed line = 90\% energy threshold. Later layers (yellow, Layer 10--11) reach the threshold at lower $k\%$, indicating higher spectral compressibility in task-specific upper layers.}
\label{fig:layer_curves}
\end{figure}

\begin{figure}[H]
\centering
\includegraphics[width=\textwidth]{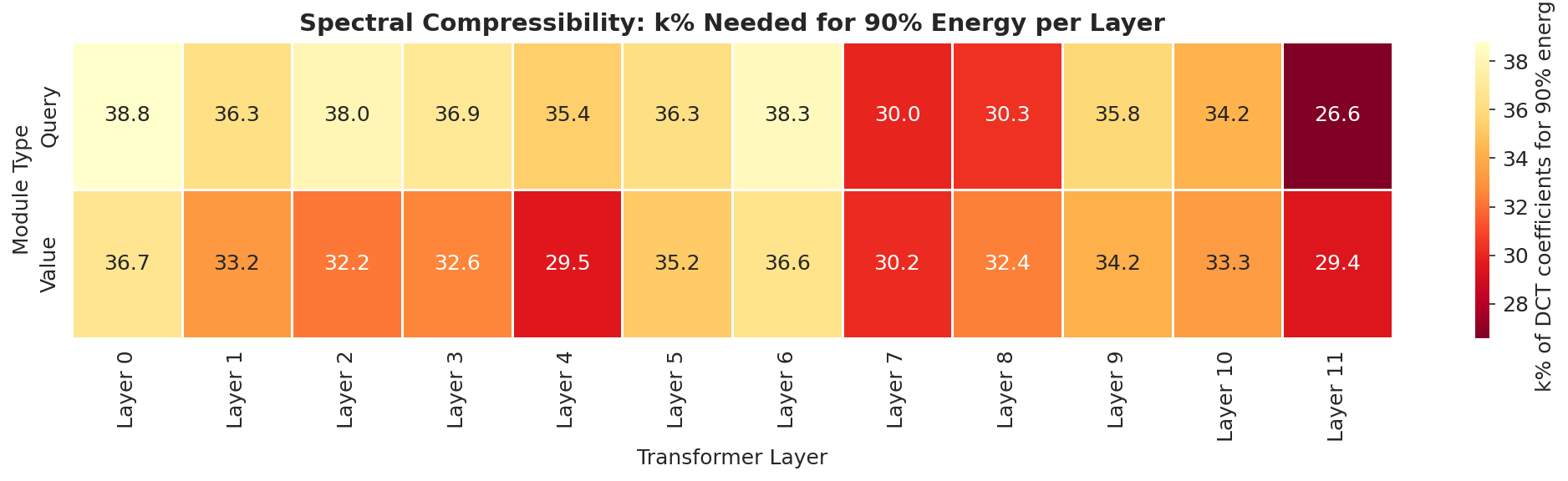}
\caption{Heatmap of $k\%$ needed for 90\% DCT energy per transformer layer and module type (Query/Value). Darker red = more compressible (lower $k\%$ needed). Layer 11 query projection requires only 26.6\% vs.\ Layer 0's 38.8\% -- a 12pp gap indicating depth-dependent spectral structure. Value projections show a different pattern, with Layer 4 being notably compressible (29.5\%).}
\label{fig:heatmap}
\end{figure}

Figure~\ref{fig:layer_curves} shows the cumulative DCT energy curves per transformer layer
for query and value projections. Figure~\ref{fig:heatmap} provides a heatmap of the $k\%$
needed for 90\% energy per layer and module type.

\subsection{Key Observations from Layer Analysis}

\begin{itemize}[leftmargin=1.5em, itemsep=3pt]
  \item \textbf{Depth gradient in query projections:} Layer 11 query requires 26.6\% vs.\
        Layer 0's 38.8\% -- a 12pp gap. Later layers adapt more smoothly.
  \item \textbf{Query vs.\ Value asymmetry:} Value projections (Layer 4: 29.5\%) show different
        compressibility patterns from query projections, suggesting different functional roles
        in spectral adaptation.
  \item \textbf{Non-monotonic value pattern:} Value projection compressibility does not
        monotonically increase with depth (Layer 4 is most compressible), unlike query projections.
  \item \textbf{Practical implication:} Layer-adaptive $k$ assignment -- smaller $k$ for
        later query layers, larger $k$ for early layers -- could recover most of the accuracy
        lost at uniform $k$=10\%.
\end{itemize}

\section{Analysis and Discussion}

\subsection{Why This Matters for PEFT Design}

Our findings collectively support a new design axis for PEFT: instead of controlling
the rank $r$ of adaptation matrices, one can control the \emph{frequency budget $k$}
of their spectral representation. This axis offers three advantages over rank:
(1) continuous compression control rather than integer rank steps,
(2) post-hoc application to any trained LoRA adapter without retraining,
and (3) natural regularization at appropriate $k$ values.
The findings motivate future work on training natively in the frequency domain. Separately, a subsequent experiment further reveals that SVD energy concentration and DCT compressibility are strongly correlated across layers (Pearson $r = 0.906$, Spearman $r = 0.935$, $p < 10^{-9}$), consistent with the Dyson Brownian Motion framework of \citet{olsen2025sgd}, which predicts that SGD creates weight matrices where energy concentrates in dominant directions. The depth-dependent pattern---later layers more compressible in both decompositions---holds consistently across both SVD and DCT analyses, suggesting SpectralLoRA's empirical $\sim 33\%$ constant may have a theoretical grounding in the stationary distribution of singular values under SGD dynamics.

\subsection{Limitations}

\begin{itemize}[leftmargin=1.5em, itemsep=2pt]
  \item Experiments use 5{,}000 training samples per task; full dataset results may
        differ quantitatively, though we expect qualitative findings to hold.
  \item Analysis focuses on attention query/value projections; FFN layers are not
        analyzed and may show different spectral structure.
  \item Results cover encoder-only models; decoder-only LLMs may exhibit different
        spectral behavior due to autoregressive training.
  \item The $\sim 33\%$ universal constant is empirical; while a connection to the Dyson Brownian Motion stationary distribution \citep{olsen2025sgd} is suggested by our SVD--DCT correlation analysis, a rigorous theoretical derivation remains future work.
\end{itemize}

\section{Conclusion}

We presented SpectralLoRA, a systematic empirical study establishing that LoRA weight
updates are universally low-frequency in the DCT domain. Our key findings are:
(1) a near-universal $\sim$33\% spectral constant for GLUE-scale adaptation,
(2) RoBERTa-base is systematically more spectrally compressible than BERT-base,
suggesting pretraining quality governs adaptation frequency structure,
(3) task complexity determines spectral sensitivity, with NLI requiring 3.7$\times$
more frequency budget than sentiment analysis, and
(4) frequency masking at $k$=50\% acts as implicit regularization, improving over
full LoRA in 3 of 8 settings.

\medskip
\noindent\colorbox{lightblue}{\parbox{0.97\textwidth}{\small\textbf{Main Takeaway:}\ LoRA weight updates are spectrally sparse. 66\% of DCT coefficients carry less than 10\% of adaptation energy. This sparsity is universal, architecture-dependent, and task-sensitive -- three properties that together define a new compression and design axis for PEFT.}}


\end{document}